\documentclass[10pt]{article}

\usepackage[preprint]{rlj}

\usepackage{amssymb}
\usepackage{mathtools}
\usepackage{mathrsfs}
\usepackage{graphicx}
\usepackage{subcaption}
\usepackage[space]{grffile}
\usepackage{url}
\usepackage{lipsum}
\usepackage{multirow}
\usepackage{wrapfig}
\usepackage{dashrule}

\title{From Demonstrations to Rewards: Test-Time Prompt Optimization for \vlm Reward Models}

\setrunningtitle{From Demonstrations to Vision-Language Rewards}

\author{Christian Gumbsch$^{1}$, Leonardo Barcellona$^{1}$, Lennard Schünemann$^{1}$, Platon Karageorgis$^{1}$, Andrii Zadaianchuk$^{1}$, Zehao Wang$^{2}$, Sergey Zakharov$^{3}$, Fabien Despinoy$^{4}$, Rahaf Aljundi$^{4}$, \& Efstratios Gavves$^{1}$}

\emails{c.gumbsch@uva.nl}

\affiliations{
$^{1}$\textbf{University of Amsterdam}\\
$^{2}$\textbf{Catholic University of Leuven}\\
$^{3}$\textbf{Toyota Research Institute}\\
$^{4}$\textbf{Toyota Motor Europe}
\par
}

\contribution{
    We propose \demoreward, a demonstration-guided test-time prompt optimization method that aligns frozen \vlm reward models for sparse binary RL without additional model training or additional computation when interacting with the environment at test-time.
    }
    {Recent work used \vlms as zero-shot reward models (\eg \citealt{RoboClip, VLM-RM, RL-VLM-F, SENSEI, WorldGym}). However, their failure conditions remain unclear.
    \demoreward leverages a small set of expert demonstrations and employs techniques from automatic prompt engineering \citep{APE_survey} to refine language instructions used by the \vlm reward model. Candidate prompts are evaluated on a set of demonstrations using an objective function that prioritizes minimizing false positives, while preserving true positives. Finally, the optimized instruction is used for policy learning by directly prompting the \vlm for binary rewards. The \vlm weights remain frozen at all time.}

\contribution{
    We provide a unified evaluation of zero- and few-shot \vlm-based reward models, within RL training schemes and across simulated robot manipulation tasks. We demonstrate that \demoreward improves policy performance and outperforms \vlm-based baselines.
    }
    {
    Using a consistent experimental setup, we observe that several existing \vlm-based reward models
    underperform when deployed directly inside RL training schemes. We hypothesize that false positive reward predictions constitute a key limiting factor. Through a case analysis, we show that high false positive rates can lead to severe reward hacking. Consequently, explicitly reducing false positives via \demoreward leads to substantial improvements in policy learning.
    }

\contribution{
    We show that \demoreward transfers to a real-world robotic learning scenario.
    }
    {
    On a physical robotic manipulation task, policies trained with \demoreward-aligned \vlm rewards achieve comparable performance than training with ground-truth rewards.
    }

\keywords{Reward Modeling, Robot Learning, Vision Language Models, Foundation Models}

\summary{
Reinforcement Learning (RL) relies on accurate reward functions, which are often handcrafted or even unavailable in real-world applications, such as robotics. Recent work has explored the zero-shot reasoning capabilities of pre-trained Vision-Language Models (\vlms) as reward models. However, without careful prompt engineering, these approaches tend to produce suboptimal rewards, where false positive predictions can severely degrade downstream policy learning. In robotics, limited datasets comprising expert demonstrations are often collected to bootstrap policy learning. This scenario provides an opportunity to optimize a reward model prior policy training.
We propose \demoreward a test-time adaptation technique to optimize the language instruction of a reward model based on a few demonstrations (3-10 trajectories) to reduce false positives while preserving true positives.
Crucially, this requires no additional model training or computation resources during policy learning.
We show that \demoreward consistently outperforms existing zero- and few-shot \vlm reward models across a range of simulated robotic tasks and policy backbones. 
Finally, we demonstrate that \demoreward effectively transfers to a real-world robotic learning scenario, enabling policy learning without manually engineering a reward function.
}

\usepackage{booktabs}

\usepackage{graphics}
\usepackage{mathtools}

\usepackage{enumitem}

\usepackage{amsfonts}
\usepackage{amsmath}
\usepackage{amssymb}
\usepackage{amsthm}
\usepackage{pifont}%
\usepackage{diagbox}
\usepackage{multirow}

\makeatletter
\newcommand\notsotiny{\@setfontsize\notsotiny{6.31415}{7.1828}}
\makeatother

\newcommand{\Fig}[1]{Figure~\ref{#1}}
\newcommand{\fig}[1]{Fig.~\ref{#1}}

\newcommand{\eqn}[1]{Eq.~\ref{#1}}

\renewcommand{\sec}[1]{Sec.~\ref{#1}}
\newcommand{\supp}[1]{Suppl.~\ref{#1}}

\usepackage{xspace}
\makeatletter
\DeclareRobustCommand\onedot{\futurelet\@let@token\@onedot}
\def\@onedot{\ifx\@let@token.\else.\null\fi\xspace}
\def\eg{e.g\onedot}
\def\ie{i.e\onedot}

\makeatother

\let\originalleft\left
\let\originalright\right
\renewcommand{\left}{\mathopen{}\mathclose\bgroup\originalleft}
\renewcommand{\right}{\aftergroup\egroup\originalright}

\newcommand{\g}[1]{%
  \ifthenelse{\equal{#1}{(}}
  {\left( }%
    { \ifthenelse{\equal{#1}{)}}
      { \right)}%
    { \ifthenelse{\equal{#1}{[}}
      {\left[}%
        { \ifthenelse{\equal{#1}{]}}
          { \right]}%
        {#1}}
    }
  }
}

\definecolor{ourblue}{rgb}{0.368,0.507,0.71}
\definecolor{ourorange}{rgb}{0.881,0.611,0.142}
\definecolor{ourgreen}{rgb}{0.56,0.692,0.195}
\definecolor{ourred}{rgb}{0.923,0.386,0.209}
\definecolor{ourviolet}{rgb}{0.528,0.471,0.701}
\definecolor{ourbrown}{rgb}{0.772,0.432,0.102}
\definecolor{ourlightblue}{rgb}{0.364,0.619,0.782}
\definecolor{ourdarkgreen}{rgb}{0.572,0.586,0.}
\definecolor{ourdarkblue}{RGB}{28,99,148}
\definecolor{ourdarkred}{RGB}{169,53,17}

\usepackage{etoolbox}
\makeatletter
\patchcmd{\NAT@test}{\else \NAT@nm}{\else \NAT@nmfmt{\NAT@nm}}{}{}
\DeclareRobustCommand\citeposs
  {\begingroup
   \let\NAT@nmfmt\NAT@posfmt%
   \NAT@swafalse\let\NAT@ctype\z@\NAT@partrue
   \@ifstar{\NAT@fulltrue\NAT@citetp}{\NAT@fullfalse\NAT@citetp}}

\let\NAT@orig@nmfmt\NAT@nmfmt
\def\NAT@posfmt#1{\NAT@orig@nmfmt{#1's}}
\makeatother

\usepackage{xr-hyper}

\makeatletter
\newcommand*{\addFileDependency}[1]{%
  \typeout{(#1)}
  \@addtofilelist{#1}
  \IfFileExists{#1}{}{\typeout{No file #1.}}
}
\makeatother

\definecolor{springreen}{RGB}{225, 247, 208}
\definecolor{outlinegreen}{RGB}{145, 170, 126}
\definecolor{boxyellow}{RGB}{245, 238, 156}
\definecolor{darkyellow}{RGB}{145, 138, 56}
\definecolor{lightpink}{RGB}{249, 203, 203}
\definecolor{gateclosed}{RGB}{107, 107, 107}
\definecolor{greyish}{RGB}{100, 100, 100}
\definecolor{pastelpurple}{RGB}{199, 206, 234}
\definecolor{darkpurple}{RGB}{99, 106, 134}
\definecolor{pastelorange}{RGB}{255, 218, 193}
\definecolor{darkorange}{RGB}{155, 118, 93}
\definecolor{pastelblue}{RGB}{192, 228, 241}
\definecolor{darkblue}{RGB}{92, 128, 141}
\definecolor{pastelred}{RGB}{255, 154, 162}
\definecolor{darkred}{RGB}{155, 54, 62} 
\definecolor{lightgrey}{RGB}{230, 230, 230}
\definecolor{wmblue}{RGB}{0, 146, 179}
\definecolor{darkwmblue}{RGB}{38, 112, 130} 
\definecolor{darkbrown}{RGB}{132, 119, 113}
\definecolor{darkwmgreen}{RGB}{93, 133, 51} 
\definecolor{gptgreen}{RGB}{12, 163, 128}
\definecolor{newpurple}{RGB}{101, 79, 167} 
\definecolor{pastelpurple}{RGB}{171,159,209}
\definecolor{lightpurple}{RGB}{169, 176, 204}
\definecolor{newlightpurple}{RGB}{169, 181, 214}
\definecolor{darknewpurple}{RGB}{50, 39, 84} 

\definecolor{midgrey}{RGB}{200, 200, 200}%
\definecolor{darkgreen}{rgb}{0.568627450980392,0.666666666666667,0.494117647058824}

\definecolor{robodeskorange}{RGB}{253, 161, 85}
\definecolor{greyish}{RGB}{100, 100, 100}

\definecolor{pink}{RGB}{255, 51, 155}
\definecolor{magenta}{RGB}{192, 37, 113}

\definecolor{darkdarkred}{rgb}{0.30392156862745096, 0.10588235294117648, 0.1215686274509802} 
\definecolor{lightdarkred}{rgb}{0.8009006473402759, 0.42851111736560654, 0.4580073177596394} 

\definecolor{lightdarkwmgreen}{rgb}{0.45588235294117646, 0.6519607843137256, 0.25}

\definecolor{darkdarkwmgreen}{rgb}{0.18235294117647058, 0.2607843137254902, 0.10000000000000003}

\newcommand{\demoreward}{Demo2Reward\xspace}

\newcommand{\llm}{\textsc{Llm}\xspace}
\newcommand{\llms}{\textsc{Llm}s\xspace}

\newcommand{\vlm}{\textsc{Vlm}\xspace}
\newcommand{\vlmsd}{\textsc{Vlm-Sd}\xspace}
\newcommand{\vlms}{\textsc{Vlm}s\xspace}

\newcommand{\gvl}{\textsc{Gvl}\xspace}

\hypersetup{
    colorlinks,
    linkcolor=newpurple,
    citecolor=newpurple,
    urlcolor=newpurple
}

\usepackage{appendix,minitoc}
\usepackage{subcaption}
\usepackage{graphicx}

\begin{document}
\dosecttoc
\faketableofcontents%

\maketitle

\begin{abstract}
Reinforcement learning relies on accurate reward functions, which are often handcrafted or even unavailable in real-world applications, such as robotics. Recent work has explored the zero-shot reasoning capabilities of pre-trained Vision-Language Models (\vlms) as reward models. However, without careful prompt engineering, these approaches tend to produce suboptimal rewards, where false positive predictions can severely degrade downstream policy learning. In robotics, limited datasets comprising expert demonstrations are often collected to bootstrap policy learning. This scenario provides an opportunity to optimize a reward model prior policy training.
We propose \demoreward a test-time adaptation technique to optimize the language instruction of a reward model based on a few demonstrations (3-10 trajectories) to reduce false positives while preserving true positives.
Crucially, this requires no additional model training or computation resources during policy learning.
We show that \demoreward consistently outperforms existing zero- and few-shot \vlm reward models across a range of simulated robotic tasks and policy backbones. 
Finally, we demonstrate that \demoreward effectively transfers to a real-world robotic learning scenario, enabling policy learning without manually engineering a reward function.
\end{abstract}

\begin{figure}
\vspace*{-0.2cm}
\centering
\begin{subfigure}{0.63\linewidth}
\centering
  \centering
  \includegraphics[width=\linewidth]{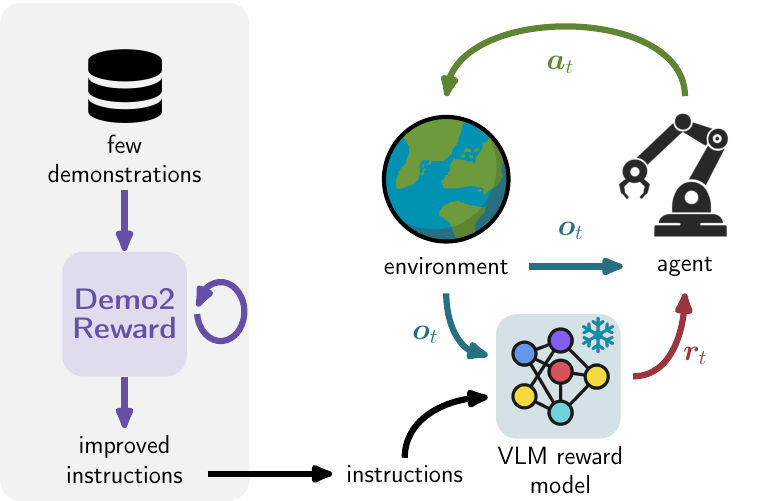}
  \caption{\textbf{\demoreward: Training Pipeline} \label{fig:overview}}
\end{subfigure}\hfill
\begin{subfigure}{0.37\linewidth}
  \centering
  \begin{tabular}{@{}cc@{}}
    \includegraphics[width=0.45\linewidth]{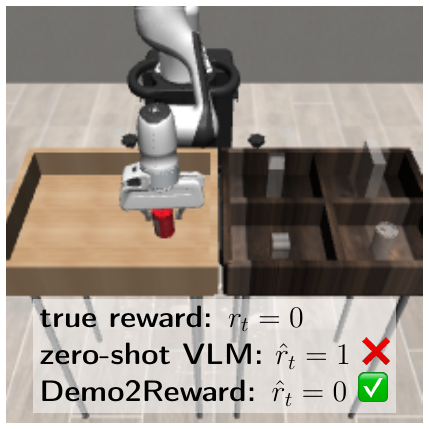} &
    \includegraphics[width=0.45\linewidth]{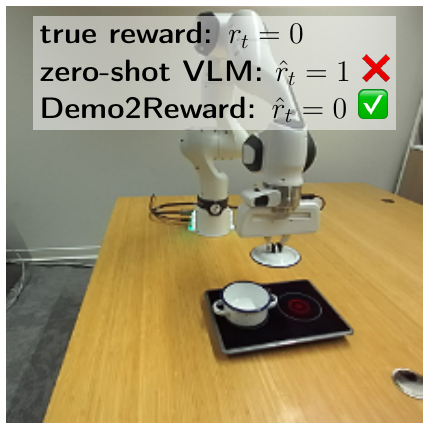} \vspace*{-0.5em}\\
    \footnotesize \texttt{PickPlace-Can} &
    \footnotesize \texttt{Lid-On-Pot} \\
    \includegraphics[width=0.45\linewidth]{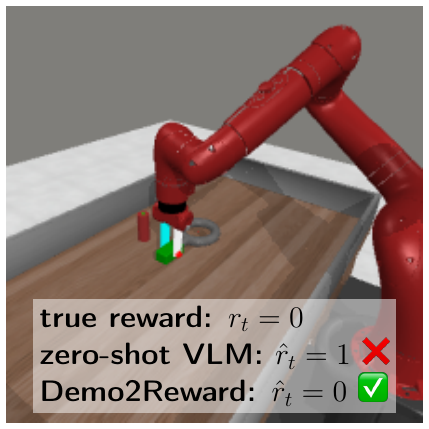} &
    \includegraphics[width=0.45\linewidth]{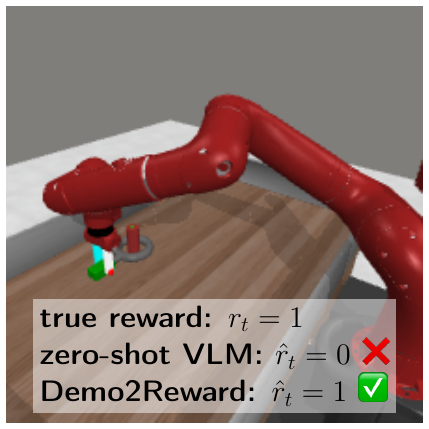} \vspace*{-0.5em}\\
    \footnotesize \texttt{Assemble-Nut} &
    \footnotesize \texttt{Assemble-Nut} 
  \end{tabular}
  \caption{\textbf{Example Rewards} \label{fig:examples}}
\end{subfigure}
\vspace*{-0.5cm}
\caption{\textbf{\demoreward overview}: \textbf{(a)} \textit{Before} policy training, \demoreward optimizes a language instruction to improve success predictions of a \vlm on a small set of demonstrations. \textit{During} policy learning, the \vlm generates rewards from video observations and the optimized instruction.  \textbf{(b)} Exemplary \vlm-generated binary rewards with final frames of videos. Without \demoreward, \vlms often prematurely detect success, \eg already during the  pick-up of pick-and-place tasks such as \texttt{Assemble-Nut} (left) or \texttt{PickPlace-Can}, while missing true successes, \eg \texttt{Assemble-Nut} (right). \demoreward yields more accurate rewards that match ground-truth success.}
\end{figure}

\section{Introduction}

Rewards are one of the main components of reinforcement learning (RL) and largely drive the policy learning. 
In the classic Markov Decision Process formulation, rewards are part of the environment and are produced by the transition function \citep{SuttonBook}. 
However, in most real-world applications, such as an RL-based robotic scenario, reward functions are not naturally available and must be manually specified.
This is a major bottleneck for embodied RL, since it requires task-specific expertise and careful reward engineering. 
In practice, reward functions frequently depend on fragile heuristic-based criteria, such as registering successful object placements by tracking colored pixels in parts of a camera image \citep{IBRL}.
As collecting robot-teleoperated demonstrations is usually simpler, state-of-the-art robotics has largely relied on supervised policy learning schemes, such as imitation learning and behavior cloning \citep{OpenVLA,Pi_05,bjorck2025gr00tn1,lingbot-va2026}. These methods are straightforward to apply, however, their performance is constrained by both the quality and quantity of available demonstrations.

Recently, pre-trained foundation models such as Large Language Models (\llms) and Vision Language Models (\vlms) have gained attention as zero-shot or few-shot reward models \citep{VLM-RM, GVL, VLM-SD, RL-VLM-F, RoboReward}. These models are trained on internet-scale data and exhibit strong capabilities in semantic interpretation, visual scene understanding, and common sense reasoning \citep{CLIP, flamingo, gemini, gpt4, Qwen3}.
Logically, such capabilities can directly be repurposed for reward modeling. Prior work applied them out of the box, \ie without fine-tuning, to generate binary task success \citep{WorldGym, WorldGymnast}, to produce continuous task progress scores \citep{GVL}, and to provide preferences in place of human annotators in preference-based RL \citep{RL-VLM-F, Motif, SENSEI}. 

Despite promising results, such an out of the box approach is extremely challenging, as even small reward misspecifications can cause the agent to exploit the reward function in unintended ways, consequently creating drastic changes in the learned policy, a phenomenon known as \textit{reward hacking} \citep{RewardHacking}. In particular, false positive rewards can lead the policy to seek out states that appear valuable to the \vlm but do not reflect the true task success, reinforcing spurious behaviors and drifting away from real task completion.

One strategy to mitigate this reward exploit is to rely on advanced prompt engineering techniques. Especially, a comprehensive understanding of the task can help to carefully design detailed prompts that clarify the task objective and its possible inner sub-goals, while reducing hallucinated rewards.
For instance, prior approaches that guide policy learning with \llms or \vlms provide explicit game rules \citep{LLMTeacher}, extensive in-context examples \citep{Voyager}, or detailed interpretations of game sprites \citep{SENSEI}. 
However, this reintroduces the original reward-specification bottleneck and requires manual design.
Conversely, automatic prompt engineering \citep{APE, APE_survey}, where one \vlm generates or refines prompts for another \vlm, represents a promising solution, but systematic evaluations of such approaches in RL settings remain limited \citep{SENSEI}.

We propose \textbf{\demoreward}, a simple and effective test-time adaptation scheme to leverage pre-trained \vlms as success detectors in sparse reward RL (illustrated in \fig{fig:overview}). 
\demoreward assumes a standard robot learning setup \citep{DDPG_from_demos, DAPG, DBAP, IBRL} in which a small number of demonstrations are collected by a human expert, before the policy optimization of a given robotic task. To generate a customized task-based instruction, \demoreward employs a \vlm as a meta-critic to optimize its own language instruction based on the provided demonstrations. \demoreward starts from a generic task instruction, and based on sampled demonstrations, the meta-critic refines the language instruction, which is then evaluated on the rest of demonstrations until false rewards are minimized.
The optimized prompt is then fixed and used during policy learning to assign rewards. 
In this way, \demoreward performs automatic prompt engineering on a small set of demonstrations to reduce false positive hallucinations of the \vlm while recognizing true successes (examples shown in \fig{fig:examples}). Importantly, this adaptation does not require additional training or fine-tuning of the \vlm. All optimization is performed \textit{before} policy learning begins, so computation and throughput during test-time, \ie when interacting with the environment during policy learning, remain unchanged. 
We show that \demoreward improves the policy learning across tasks, simulation and real-world environments, as well as RL backbones.

Our main contributions are as follows:
\begin{enumerate}
    \item We propose \demoreward, an automatic prompt engineering scheme that improves pre-trained \vlm reward models without additional training or test-time compute.
    \item We analyze various \vlm-based reward models for policy optimization in simulated robot tasks and show that \demoreward mitigates these issues and improves policy learning across environments, policy backbones, and generated prompts.
    \item We demonstrate that \demoreward transfers to real-world robot learning and enables RL without manually specifying a reward function.
\end{enumerate}

\section{Related Work}

\textbf{\vlms as source of rewards:} There is an expanding number of work that leverages pre-trained foundation models, such as \llms and \vlms, to generate rewards for RL agents. These approaches can be grouped into  \textbf{(1) explicit prompting}, \textbf{(2) preference-based methods}, \textbf{(3) latent embedding approaches}, and \textbf{(4) code generation}. \textbf{Explicit prompting} directly queries a foundation model to produce scalar rewards. The simplest variant predicts a binary task success, as in SuccessVQA \citep{VLM-SD}. Similarly, WorldGym \citep{WorldGym} uses a frozen \vlm to evaluate imagined rollouts within a video diffusion model to train policies in simulation \citep{WorldGymnast}. Beyond binary success, \vlms are also prompted to provide discrete or continuous task progress rewards \citep{RoboReward, RoboDopamine, GVL}. For example, RoboReward \citep{RoboReward} and Robo-Dopamine \citep{RoboDopamine} fine-tune \vlms to assign discrete task process rewards to robotic videos, while \textsc{Gvl} \citep{GVL} demonstrates that even frozen \vlms can estimate task completion percentages with a suitable prompting scheme.
\textbf{Preference-based methods} instead distill a reward model from \vlm-generated preferences, following the paradigm of RL from human feedback \citep{RLHF}. This is applied to text-based environments using \llm annotators \citep{Motif}, to visual observations using \vlm annotators \citep{RL-VLM-F, SENSEI}, and to multimodal settings by combining \llm and \vlm annotations \citep{PRIMT}. 
\textbf{Latent embedding approaches} derive synthetic rewards by comparing \vlm embeddings of observations and task descriptions, for example using \textsc{Clip} similarities as in RoboCLIP \citep{RoboClip} and \textsc{Vlm-RM} \citep{VLM-RM}. 
Finally, \textbf{code generation methods} generate executable reward functions from task descriptions \citep{l2r, text2reward, VLMCar}. In this work, we focus on explicit prompting approaches and their optimization at test time.

\textbf{Automatic Prompt-Engineering (APE)} refers to methods that automatically refine the text-based input to a language model in order to improve performance with respect to a given objective. Various techniques are proposed to optimize prompts, including refinement through a meta-model, gradient-based optimization, genetic algorithms, and reinforcement learning \citep{APE_survey}. APE is successfully applied to natural language reasoning \citep{APE}, code generation \citep{CEDAR}, and multimodal generation \citep{Promptist}. However, its usage to learn reliable reward models in reinforcement learning remains limited. \textsc{Sensei} \citep{SENSEI} shows that prompts generated by a \vlm from initial screenshots can yield better rewards for task-free exploration than manually designed prompts.
To the best of our knowledge, APE for test-time optimization of \vlm reward models has not yet been explored.

\textbf{In-context learning (\textsc{Icl})} is an alternative technique to APE for adapting \llms or \vlms at test time. In \textsc{Icl}, example input-output pairs are provided as part of the prompt to allow the model to infer the underlying pattern and generate outputs accordingly. For reward modeling, \textsc{Icl} is primarily studied in the context of improving preference-based RL \citep{ICPL}. For explicit prompting approaches, \textsc{Gvl} \citep{GVL} proposes a prompting scheme and shows that in-context demonstrations can improve value and task progress predictions.

\textbf{Inverse reinforcement learning (\textsc{Irl})} is a classical approach for learning reward functions from expert demonstrations \citep{abbeel2004, ziebart2008}. \textsc{Irl} infers a reward under which the demonstrations are approximately optimal and typically requires repeated policy optimization during training, as well as tens to hundreds of expert trajectories to obtain stable reward estimates \citep{ziebart2008, Wulfmeier2015, ho2016}. In contrast, \demoreward does not attempt to recover a reward function from scratch. Instead, it leverages a pre-trained \vlm and adapts its prompt using only a handful of demonstrations before policy learning begins.

\section{Method}

\subsection{Problem Setup}

We consider tasks that can be formalized as a partially observable Markov decision process (POMDP) $\mathcal{M} = (\mathcal{S}, \mathcal{O}, \mathcal{A}, T, R, \ell)$ \citep{kaelbling1998} with state space $\mathcal{S}$, observation space $\mathcal{O}$, action space $\mathcal{A}$, transition function $T: \mathcal{S} \times \mathcal{A} \rightarrow \mathcal{S} \times \mathcal{O}$ and a binary success/no-success reward function $R: \mathcal{S} \rightarrow \{0, 1\}$. The task is specified by a natural language instruction $\ell$.

At each time step $t$ the agent receives an observation $\mathbf{o}_t \in \mathcal{O}$, which consists of an RGB image and may additionally include proprioceptive inputs such as robot joint angles.
The true state $\mathbf{s}_t \in \mathcal{S}$ is not observable.
We further assume that the reward is not directly accessible to the agent. Instead, the agent must approximate the true reward $r_t$ with a synthetic reward $\hat{r}_t$ that is used to train a policy.

As is common in robotics \citep{DDPG_from_demos, IBRL}, we assume access to a small dataset of expert demonstrations $\mathcal{D} = \left\{ \tau^i \right\}^N_{i=1}$ with $3 \leq N \leq 20$. Each trajectory is given by $\tau^i = \big( (\mathbf{o}_1, \mathbf{a}_t, r_1), \dots, (\mathbf{o}_T, \mathbf{a}_T, r_T) \big)$.
The demonstrations are collected either through direct human teleoperation in the real world or from a near-optimal policy in simulation. 
Since rewards are not observable from the environment, for each data point $(\mathbf{o}, \mathbf{a}_t, r_t) \in \mathcal{D}$ a reward label $r_t$ is provided by a human expert and indicates task success. This dataset can be used to initialize a policy via behavioral cloning, but more importantly it serves as supervision for reward modeling.

\subsection{Motivating Example: Reward Hacking with \vlms}
\label{sec:reward_hacking}

Since the agent cannot access the true reward $r_t$, it requires a reward model $m_\mathrm{critic}$ that produces synthetic rewards $\hat{r}_t$.
Following recent approaches that leverage a pre-trained \vlm as a zero-shot reward model (\eg \citealt{WorldGym, WorldGymnast, RoboReward}), we define:
\begin{equation}
\hat{r}_t = m_{\mathrm{critic}}(\mathbf{o}_{1:t}, \ell)
\end{equation}
as the output of the \vlm given observations $\mathbf{o}_{1:t}$, a task instruction $\ell$ and a generic  template that instructs the \vlm to output a binary scalar reward (prompt details provided in \supp{supp:vlmsd}).

\begin{figure}
    \centering
\begin{subfigure}{0.74\linewidth}
\centering
\includegraphics[width=0.87\linewidth]{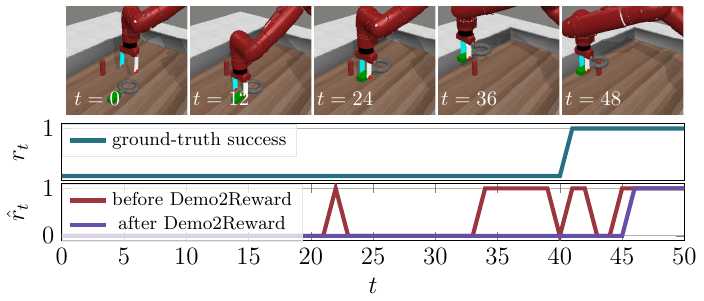}
\caption{ \textbf{Reward predictions for \texttt{Assemble-Nut}} \label{fig:rewardhack_traj}}
\end{subfigure}
\hfill
\begin{subfigure}{0.25\linewidth}
\centering
\includegraphics[width=\linewidth]{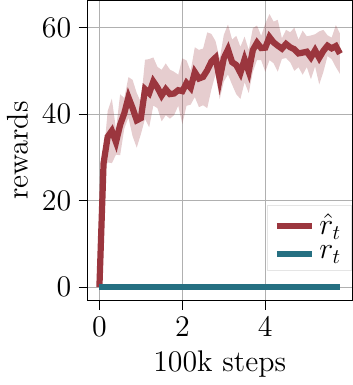}
\caption{\textbf{Reward collection \label{fig:rewardhack_rew}}}
\end{subfigure}
    \caption{\textbf{Reward Misalignment with \vlms}:  Without guidance, a \vlm reward model can easily misjudge rewards.  \textbf{(a)} In an example \texttt{Assemble-Nut} sequence the \vlm-generated rewards output false positives ($\hat{r}_t$ before Demo2Reward, bottom row) before the the task is actually complete (ground-truth reward $r_t$, top row).  \textbf{(b)} Training an agent on these flawed rewards causes it to exploit such false positives. The agent collects high synthetic rewards ($\hat{r}_t$, \textcolor{darkred}{red}) without ever reaching true task success ($r_t=1$, \textcolor{darkwmblue}{blue}). Shaded area shows standard deviation (5 seeds). } \label{fig:rewardhack}
\end{figure}

Exploiting the zero-shot adaptability of \vlms, we train a policy solely using rewards from $m_{\mathrm{critic}}$ (details in \supp{supp:sim_details}). As illustrated in \Fig{fig:rewardhack}, although the predicted reward $\hat{r}_t$ increases during training, true task success is never reached (\fig{fig:rewardhack_rew}). By comparing the true reward from the environment with the predicted reward over time (\fig{fig:rewardhack_traj}), we observe that the latter produces many false-positive signals. This high frequency of false positives effectively induces severe reward hacking, resulting in a policy that is unable to solve the task.

\subsection{\demoreward}

To address the false positive reward issue, we introduce \demoreward, which adapts the task instruction of a pre-trained \vlm using a small dataset of demonstrations before the policy learning begins.
Instead of directly using the task instruction $\ell$, we search for an improved instruction $p$ that yields more reliable reward predictions on the demonstrations. Since we only have few demonstration of the task, we extend the dataset $\mathcal{D}$ by generating additional samples from each trajectory. 
Specifically, for each demonstration of length $T$, we construct $T$ prefix subsequences 
$(o_1, r_1), \; (o_1, r_1, o_2, r_2), \; \dots, \; (o_1, r_1, \dots, o_T, r_T).$
This results in an augmented dataset $\mathcal{D}_{\mathrm{sub}}$ containing $N \times T$ clips for $N$ original demonstrations.

Next, to optimize the instruction on our subsampled dataset $\mathcal{D}_\mathrm{sub}$, we introduce a second \vlm  $m_{\mathrm{meta}}$, referred to as the meta-critic. 
The meta-critic receives a sampled demonstration, the current instruction $p$, and representative false annotations for the instruction $p$, and proposes an updated instruction $p'$ (for the prompt pattern see \supp{supp:meta-critic}).
We evaluate the critic model with the new prompt containing $p'$ on $\mathcal{D}_{\mathrm{sub}}$ using an objective $\widehat{\mathcal{O}}(p')$. 
If the objective improves, we replace $p \leftarrow p'$, otherwise we keep $p$.
This process is repeated for a fixed number of steps ($I=100$).

\paragraph{Objective} Our goal is to maximize correct reward predictions for a task instruction $p$ on the dataset:
\begin{equation}
\mathcal{O}(p)
=
\mathbb{E}_{(\mathbf{o}_{1:t}, r_t)\sim\mathcal{D}_{\mathrm{sub}}}
\Big[
(1-r_t)\big(1-m_{\mathrm{critic}}(\mathbf{o}_{1:t},p)\big)
+
r_t\, m_{\mathrm{critic}}(\mathbf{o}_{1:t},p)
\Big].
\end{equation}
In practice, we compute the true negative rate (TNR) and true positive rate (TPR), with
\begin{equation}
\mathrm{TNR}(p)
=
\frac{1}{|\mathcal D^-|}
\sum_{(\mathbf{o}_t, r_t)\in\mathcal D^-}
\mathbf{1}\{m_{\mathrm{critic}}(\mathbf{o}_{1:t},p)=0\},
\end{equation}
\begin{equation}
\mathrm{TPR}(p)
=
\frac{1}{|\mathcal D^+|}
\sum_{(\mathbf{o}_t,r_t)\in\mathcal D^+}
\mathbf{1}\{m_{\mathrm{critic}}(\mathbf{o}_{1:t},p)=1\}.
\end{equation}
Here, $\mathcal{D}^+$ and $\mathcal{D}^-$ denote the subsets of $\mathcal{D}_{\mathrm{sub}}$ with $r_t = 1$ and $r_t = 0$, respectively. We use a weighted objective of the two terms
\begin{equation}
\widehat{\mathcal{O}}(p)
=
\mathrm{TNR}(p)
+
\lambda\,\mathrm{TPR}(p). \label{eq:O}
\end{equation}
We choose a small $\lambda = 0.01$ to mainly prioritize minimizing false positives.

\paragraph{Avoiding Degenerate Solutions} Strongly weighting true negatives introduces a trivial local optimum in which 
the prompt always predicts zero reward. In practice, once such a prompt is discovered, optimization rarely recovers. To mitigate this issue, we repeat the optimization process $K$ times with  different initializations and different \vlms as meta-critic models $m_\mathrm{meta}$. 
We select the task instruction with the highest objective value.

\section{Experiments}

In our experiments, we empirically evaluate the following research questions:
\begin{enumerate}
    \item Can \demoreward discover improved instructions through its automatic prompt engineering? 
    \item Do these instructions improve the downstream policy learning when used to generate rewards?
    \item How does \demoreward compare to other zero- or few-shot \vlm-based reward models?
    \item How robust is \demoreward across different settings?
    \item Can \demoreward be applied to real-world robot learning?
\end{enumerate}

We first illustrate instruction discovery with \demoreward (\sec{sec:res1}) and show how improved instructions translate into better downstream policy learning (\sec{sec:res1b}). 
We then benchmark \demoreward against various \vlm-based reward models (\sec{sec:res1c}) and test \demoreward's robustness with respect to different RL backbones and randomly discovered instructions (\sec{sec:res2}).
Finally, we demonstrate that \demoreward enables learning a real-world robotic policy using online rewards generated solely by a \vlm (\sec{sec:res3}).

\subsection{Experimental Setup}

\paragraph{RL-based policy learning.}

In the experiments, rewards for policy training are generated using a pre-trained \vlm. Following prior work on \vlm-based reward modeling \citep{RoboClip, RoboReward}, rewards are provided only at the end of each episode. This creates a challenging setup with sparse and potentially delayed rewards, but significantly reduces computational overhead since the \vlm is queried only once per episode.
For policy learning, we use the following two algorithms that leverage demonstrations to accelerate training.

\textbf{Imitation Bootstrapped Reinforcement Learning (IBRL).}
\textsc{Ibrl} \citep{IBRL} first pre-trains an imitation learning policy on expert demonstrations, before training a second RL-based policy during online interaction.
At each time step, the agent selects the action from the policy with the highest estimated Q-value. We initialize \textsc{Ibrl} using the imitation policies provided by the authors.

\textbf{Reinforcement Learning from Prior Data (\textsc{Rlpd}).}
\textsc{Rlpd} \citep{RLPD} maintains two replay buffers, one for demonstration data and one for online interaction data. 
During training, demonstration samples are oversampled by drawing 50\% of each batch from each buffer.

We consider the following environments  and tasks.

\textbf{MetaWorld} \citep{metaworld} is an open-source benchmark featuring a Sawyer robot interacting with tabletop objects. 
Following \citep{IBRL}, we evaluate four tasks of varying difficulty: 
\texttt{Assemble-Nut}, \texttt{Box-Close}, \texttt{Coffee-Push}, and \texttt{Stick-Pull}. 
We use pixel-based observations only and provide the task descriptions from the official MetaWorld documentation as language instructions $\ell$.%
Since human teleoperation is not available for MetaWorld, demonstrations are generated using scripted controllers from the official benchmark \citep{metaworld, IBRL}.

\textbf{RoboMimic} \citep{RoboMimic} is a benchmark suite built on RoboSuite \citep{Robosuite} with human teleoperated demonstrations. 
We evaluate two object manipulation tasks with a Franka Emika Panda: \texttt{PickPlace-Can} and \texttt{NutAssembly-Square}. 
We use pixel-based observations and the official task descriptions from RoboSuite \citep{Robosuite}. 

\textbf{Real Robot}
We evaluate \demoreward on a Franka Panda robot. 
In the \texttt{Lid-On-Pot} task, the robot must close a pot with a lid. 
We collect 20 demonstrations via teleoperation. 
Observations consist of front-view RGB images (see \fig{fig:examples}) and the end-effector position (details in \supp{supp:robot}).

We compare \demoreward against several zero-shot and few-shot \vlm-based reward models. 
Unless otherwise stated, we use Qwen3-VL \citep{Qwen3} as the \vlm backbone for fair comparison.
Implementation details for all methods can be found in \supp{supp:impl_details}.

\textbf{\vlm Success Detector (\vlmsd).}
Our primary baseline uses a \vlm as a zero-shot success detector following \cite{WorldGym}. 
The model receives a subsampled video of an episode and the task instruction, and outputs a binary success label. 
We use \texttt{Qwen3-VL-8B}, which provided strong performance and higher throughput than larger models in preliminary experiments (see \supp{supp:ablation}). 

\textbf{RoboReward} \citep{RoboReward}. 
RoboReward fine-tunes \texttt{Qwen3-VL-8B} on large-scale robotics data to serve as a zero-shot foundational reward model. 
It outputs a discrete score in $\{1,2,3,4,5\}$ describing task success for a video depicting robot behavior. 
We normalize these scores to $[0,1]$.

\textbf{Generative Value Learning (GVL)} \citep{GVL}. 
\gvl uses in-context learning for value estimation with pre-trained \vlms. 
The \vlm receives a demonstration video and an interaction video and outputs a task completion percentage per frame. 
The frames within each video are randomly shuffled, forcing the \vlm to attend to the content of each frame rather than rely on temporal ordering.
We use the percentage for the final frame as the reward and normalize it to $[0,1]$. 
For \gvl, we use \texttt{Qwen3-VL-32B}, as smaller models struggled with the longer context required by this method.

\textbf{RoboCLIP} \citep{RoboClip}. 
RoboCLIP embeds demonstration and interaction videos into a latent space of a pre-trained \vlm \citep{S3D} and uses the scalar product between the latent embeddings to estimate rewards. 
We normalize rewards to $[0,1]$ for consistency.

\subsection{Instruction Discovery with \demoreward}

\label{sec:res1}

\begin{figure}
    \centering
\begin{subfigure}{0.64\linewidth}
\centering
\includegraphics[width=\linewidth]{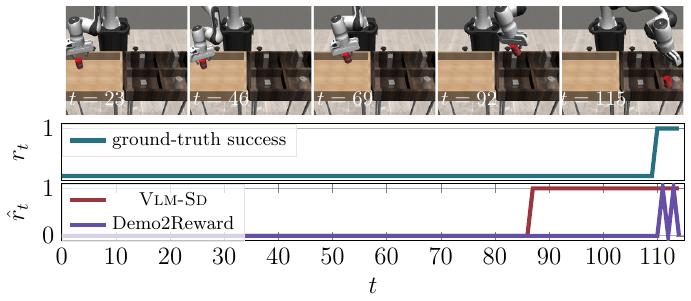}
\caption{ \textbf{Reward predictions for \texttt{PickPlace-Can}}\label{fig:example_can} }
\end{subfigure}
\hfill
\begin{subfigure}{0.35\linewidth}
\centering
\includegraphics[width=\linewidth]{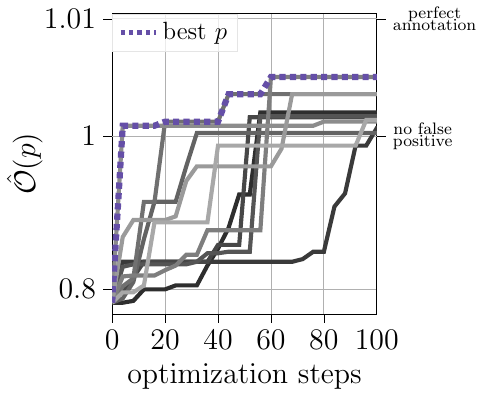}
\caption{\textbf{Optimization in \demoreward} \label{fig:O}}
\end{subfigure}
    \caption{\textbf{Instruction optimization}: \textbf{(a)} Example reward prediction for \texttt{PickPlace-Can} before (\textcolor{darkred}{red}) and after (\textcolor{newpurple}{purple}) \demoreward. \textbf{(b)} Objective $\hat{\mathcal{O}}(p)$   over optimization steps ($y$-axis in log-scale). Each gray line shows one of $K=10$ independent optimization runs with \textcolor{newpurple}{purple} marking the best. $\hat{\mathcal{O}}(p)=1.01$ marks the perfect score with 100\% TNR and 100\% TPR ($\lambda=0.01$ in \eqn{eq:O}). \label{fig:opti}}
    \vspace*{-1em}
\end{figure}

\begin{figure}[bp]
\centering
\begin{minipage}{0.4\linewidth}
\centering
\captionof{table}{Confusion matrix for $\mathcal{D}_\mathrm{sub}$ in \texttt{PickPlace-Can} \textit{before} optimization}
\vspace*{-.5em}
\begin{tabular}{c|cc}\label{tab:before}
 & \multicolumn{2}{c}{\textbf{Pred. Reward} $\hat{r}_t$} \\
\textbf{Success} $r_t$ & 1 & 0 \\ \hline
1 & 100$\%$ & 0$\%$ \\
0 & 26.1$\%$ & 73.9$\%$ \\
\end{tabular}
\end{minipage}
\hfill
\begin{minipage}{0.4\linewidth}
\centering
\captionof{table}{Confusion matrix for $\mathcal{D}_\mathrm{sub}$ in \texttt{PickPlace-Can} \textit{after} optimization} \label{tab:after}
\vspace*{-.5em}
\begin{tabular}{c|cc}
 & \multicolumn{2}{c}{\textbf{Pred. Reward} $\hat{r}_t$} \\
\textbf{Success} $r_t$ & 1 & 0 \\ \hline
1 & 68.8$\%$ & 31.2$\%$ \\
0 & 0$\%$ & 100$\%$ \\
\end{tabular}
\end{minipage}
\end{figure}

We first evaluate language instruction optimization with \demoreward on MetaWorld and RoboMimic tasks. 
For each task, we optimize the instruction using a dataset of three expert demonstrations. 
Figure~\ref{fig:opti} and Tables~\ref{tab:before}-\ref{tab:after} illustrate the optimization process for two representative tasks.

\paragraph{Instruction optimization.} The original and generic task instruction, without \demoreward, exhibits a high false positive rate (Table~\ref{tab:before}).
In particular, the \vlm tends to assign rewards before the task is actually completed (\eg \fig{fig:rewardhack_traj} \& \fig{fig:example_can}). 
During optimization, \demoreward explores alternative instruction formulations and progressively improves the objective $\hat{\mathcal{O}}$ (\fig{fig:O}). 
The final discovered instruction reduces the false positive rate to zero while maintaining a reasonable true positive rate (Table~\ref{tab:after}).
We provide more examples in \supp{supp:examples}.

\paragraph{Resulting instructions.} The full list of discovered instructions can be found in \supp{supp:prompts}. Qualitatively, discovered instructions specify visual properties of objects (\eg ``\dots {\ttfamily the nut (a gray circular object)} \dots '), emphasize spatial relations between object (\eg ``\dots  {\ttfamily the white mug is fully under the coffee machine} \dots ''), and list exact visually verifiable success condition, failure checks and ignorable aspects (\eg ``\dots  {\ttfamily Ignore slight rotation or minor misalignment} \dots ''). 
Interestingly, some discovered instructions describe visual patterns that do not strictly correspond to true elements of the scene but instead reflect how the \vlm internally represents the scene. 
For example, a blue stick may be described as a ``{\ttfamily red stick}'' due to visual artifacts from the gripper alignment. 
Although such descriptions may appear counterintuitive, they better align with the model’s perceptual biases, as they represent the \vlms's own understanding. Thus, such instructions can produce more reliable reward predictions, as reflected in the perfect true negative rate achieved by all optimized instructions in the demonstration data set.

\subsection{Downstream policy learning with \demoreward}
\label{sec:res1b}

\begin{figure}
    \centering
    {\scriptsize
      \textcolor{darkwmblue}{\rule[2pt]{20pt}{2pt}} GT Rewards \quad  
      \textcolor{darkred}{\rule[2pt]{20pt}{2pt}} \vlmsd \quad \textcolor{newpurple}{\rule[2pt]{20pt}{2pt}} \demoreward
    }\\
    \vspace{.3em}
\begin{subfigure}{0.45\linewidth}
\centering
\includegraphics[width=0.9\linewidth]{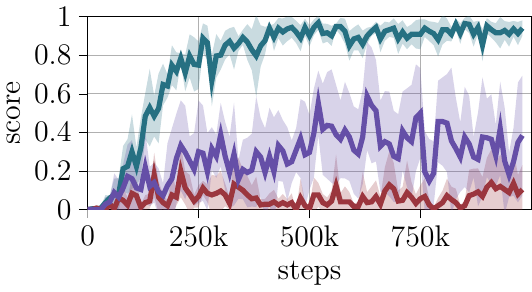}
\caption{\texttt{PickPlace-Can}}
\end{subfigure}
\hfill
\begin{subfigure}{0.45\linewidth}
\centering
\includegraphics[width=0.9\linewidth]{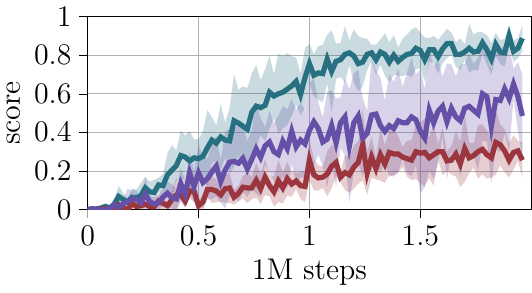}
\caption{\texttt{NutAssembly-Square}}
\end{subfigure}
\vspace*{-.75em}
    \caption{\textbf{Downstream policy learning in RoboMimic} when training with \textsc{Ibrl}. Ground-truth rewards (\textcolor{darkwmblue}{blue}) serve as an approximate upper bound. \demoreward (\textcolor{newpurple}{purple}) clearly outperforms its counterpart without optimization (\vlmsd, \textcolor{darkred}{red}). Shaded areas show standard deviation (5 seeds)\label{fig:res_robomimic}}
\end{figure}

Next, we evaluate whether improved task instructions translate into better downstream policy learning with \textsc{Ibrl} \citep{IBRL} on RoboMimic. 
Figure~\ref{fig:res_robomimic} shows the scores during the evaluation of a policy trained with synthetic rewards from a \vlm success detector (\vlmsd), \demoreward, or ground-truth rewards. 
The policy trained with \vlmsd struggles to achieve high task success, especially for the \texttt{PickPlace-Can} task.
\demoreward consistently improves upon the zero-shot baseline and reaches higher task success rates.
Thus, test-time optimization through \demoreward directly results in better downstream task learning when using \vlm-generated rewards.

\begin{figure}
    \centering
    {\scriptsize
      \textcolor{darkwmblue}{\rule[2pt]{20pt}{2pt}} GT Rewards \quad \textcolor{darkwmgreen}{\rule[2pt]{20pt}{2pt}} RoboReward \quad
      \textcolor{greyish}{\rule[2pt]{20pt}{2pt}} \gvl \quad 
      \textcolor{darkbrown}{\rule[2pt]{20pt}{2pt}} RoboCLIP \quad 
      \textcolor{darkred}{\rule[2pt]{20pt}{2pt}} \vlmsd \quad \textcolor{newpurple}{\rule[2pt]{20pt}{2pt}} \demoreward
    }\\
    \vspace{.3em}
\begin{subfigure}{0.24\linewidth}
\centering
\includegraphics[width=\linewidth]{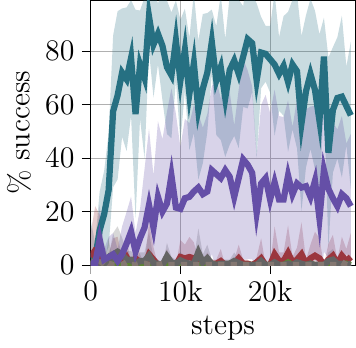}
\caption{\texttt{Assemble-Nut}}
\end{subfigure}
\hfill
\begin{subfigure}{0.24\linewidth}
\centering
\includegraphics[width=\linewidth]{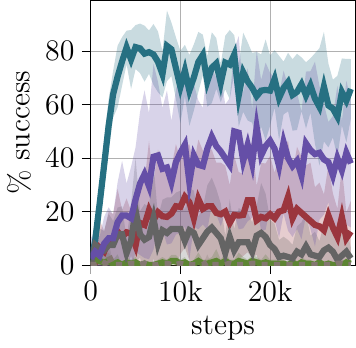}
\caption{\texttt{Box-Close}}
\end{subfigure}
\hfill
\begin{subfigure}{0.24\linewidth}
\centering
\includegraphics[width=\linewidth]{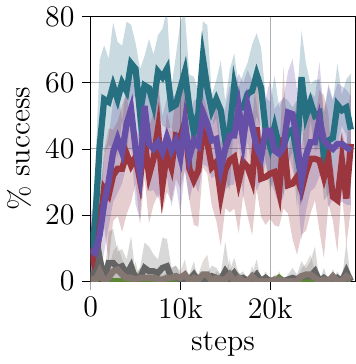}
\caption{\texttt{Coffee-Push}}
\end{subfigure}
\hfill
\begin{subfigure}{0.24\linewidth}
\centering
\includegraphics[width=\linewidth]{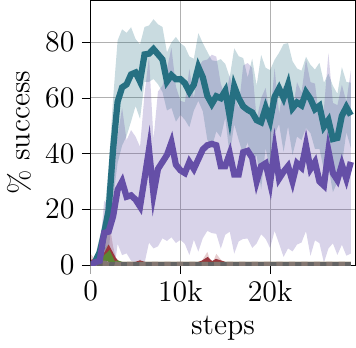}
\caption{\texttt{Stick-Pull}}
\end{subfigure}
\vspace*{-.5em}
    \caption{\textbf{Downstream policy learning in MetaWorld} by training an \textsc{Ibrl} policy with different reward sources. \demoreward  (\textcolor{newpurple}{purple}) outperforms all \vlm-based baselines (10 seeds).} \label{fig:res_metaworld}
\end{figure}

\subsection{Benchmarking \demoreward }
\label{sec:res1c}
We compare \demoreward with various state-of-the art \vlm-based reward models for policy learning  with \textsc{Ibrl} in MetaWorld.
Figure~\ref{fig:res_metaworld} reports success rates for the considered \vlm-based reward models and ground-truth rewards. 
The setup is challenging, as even with ground-truth rewards the policy peaks around 60-80$\%$. After peak success, the performance slightly declines, likely due to instabilities in sparse reward off-policy learning \citep{Discor}.

Most \vlm-based reward models fail to achieve meaningful success rates. 
Among the baselines, only \vlmsd and \gvl solve some tasks with non-zero performance, albeit substantially below ground-truth rewards.
With \demoreward, policy learning improves significantly across tasks, yielding up to a 40\% increase in success rate. 
On some tasks, \eg \texttt{Coffee-Push}, performance approaches ground-truth levels despite the absence of any ground-truth rewards during online interaction.

\subsection{Robustness of \demoreward}

\label{sec:res2}

\begin{figure}[b]
    \centering
    {\scriptsize
      \textcolor{darkwmblue}{\rule[2pt]{20pt}{2pt}} GT Rewards \quad
      \textcolor{darkred}{\rule[2pt]{20pt}{2pt}} \vlmsd \quad \textcolor{newpurple}{\rule[2pt]{20pt}{2pt}} \demoreward $p_1$ \quad
      \textcolor{darknewpurple}{\rule[2pt]{20pt}{2pt}} \demoreward $p_2$ \quad
      \textcolor{lightpurple}{\rule[2pt]{20pt}{2pt}} \demoreward $p_3$
    }\\
    \vspace{.3em}
\begin{subfigure}{0.24\linewidth}
\centering
\includegraphics[width=\linewidth]{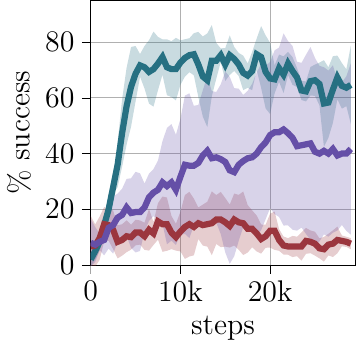}
\caption{\footnotesize  \textsc{Rlpd}: \texttt{Box-Close} \label{fig:res_rlpd_1}}
\end{subfigure}
\hfill
\begin{subfigure}{0.24\linewidth}
\centering
\includegraphics[width=\linewidth]{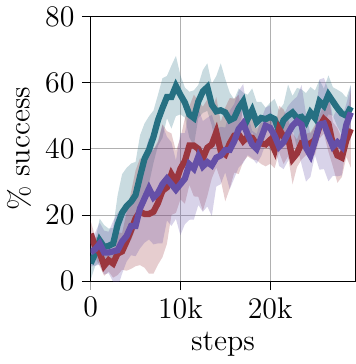}
\caption{\footnotesize \textsc{Rlpd}: \texttt{Coffee-Push} \label{fig:res_rlpd_2}}
\end{subfigure}
\hfill
\begin{subfigure}{0.24\linewidth}
\centering
\includegraphics[width=\linewidth]{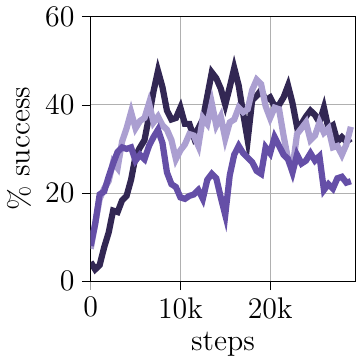}
\caption{\footnotesize \textsc{Ibrl} \texttt{Box-Close} \label{fig:res_prompt_1}}
\end{subfigure}
\hfill
\begin{subfigure}{0.24\linewidth}
\centering
\includegraphics[width=\linewidth]{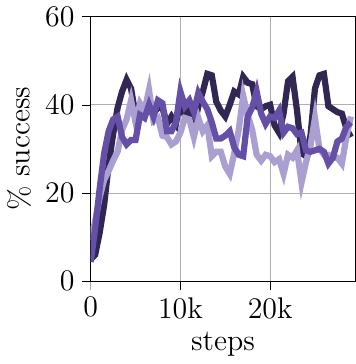}
\caption{\footnotesize \textsc{Ibrl} \texttt{Coffee-Push} \label{fig:res_prompt_2}}
\end{subfigure}
\vspace*{-.5em}
    \caption{\textbf{Robustness in MetaWorld}: We analyze the robustness of \demoreward by \textbf{(a)} applying it to a different RL backbone \textsc{Rlpd} and \textbf{(b)} comparing different discovered instructions $p$ with \textsc{Ibrl} policy learning. Results are averaged  (5 seeds) and smoothed with a sliding window
    (size $=3$).} \label{fig:robustness}
\end{figure}

\paragraph{Policy Backbones.}
To demonstrate that performance gains stem from improved reward modeling rather than a specific policy learning algorithm, we evaluate the same optimized prompts using \textsc{Rlpd} \citep{RLPD}. 
Figures~\ref{fig:res_rlpd_1}-\ref{fig:res_rlpd_2} report results for two representative MetaWorld tasks. 
When applied to policy learning with \textsc{Rlpd}, \demoreward again consistently matches or outperforms the zero-shot \vlmsd baseline.
This indicates that the gains generalize across policy backbones.

\paragraph{Instruction Variability.}
Since \demoreward performs \vlm-guided prompt optimization via sampling, different runs will produce different instructions. 
We therefore repeat instruction discovery with different random initializations for two MetaWorld tasks and evaluate policy learning with \textsc{Ibrl}. 
Figures~\ref{fig:res_prompt_1}-\ref{fig:res_prompt_2} show the resulting success rates. 
Although there is some variation between prompts, final task performance consistently reaches similar levels. 
This suggests that \demoreward can discover multiple effective instructions for stable reward generation.

\subsection{Real-World Experiments}
\label{sec:res3}

\begin{wrapfigure}[15]{r}{0.33\linewidth}
\centering

\includegraphics[width=.9\linewidth]{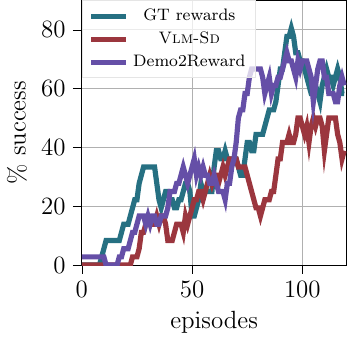}
    \caption{\textbf{Real-Robot Results} Mean success rates (3 seeds) smoothed (window size 12). \label{fig:res_real}}
\end{wrapfigure}
Finally, we evaluate real-world scalability by applying \demoreward to our \texttt{Lid-On-Pot} task on a physical robot. 
We train policies using \textsc{Rlpd} \citep{RLPD}, using the RLInf \citep{RLInf} implementation for real-world robotics, and compare ground-truth rewards from a manually defined task success criterion to zero-shot \vlmsd rewards and \demoreward.

Figure~\ref{fig:res_real} shows the resulting behaviors. 
Despite never accessing ground-truth rewards during training, the policy trained with \demoreward achieves success rates on par with using ground-truth rewards. 
In contrast, zero-shot \vlmsd suffers from frequent false positives, 
and learns a substantially weaker policy.
After around 100 episodes, \demoreward achieves a 20$\%$ higher success rate compared to vanilla \vlmsd.
These results demonstrate that \demoreward is not limited to simulated benchmarks but can successfully transfer to real-world robot learning.

\section{Discussion}

In this work, we argued and showed that naively deploying pre-trained \vlms as reward models can lead to severe reward hacking and suboptimal policy optimization. 
To address this, we have introduced \demoreward which adapts the task instruction of a frozen \vlm  based on a few demonstrations before policy learning begins.
Across simulated benchmarks and a real-world robot experiment, \demoreward consistently improved downstream policy performance without additional model training or test-time computation during policy learning. These results suggest that aligning foundation models through demonstration-guided  optimization of task instructions is a practical and scalable direction for reward modeling in embodied RL.

\paragraph{Limitations and Future Work.}
If visual conditions change between demonstration collection and policy learning, \demoreward could overfit to the small demonstration dataset. 
This could be mitigated by incorporating demonstrations from multiple camera viewpoints or by regularizing the optimization of task instructions toward the original generic ones to preserve semantic consistency. 
Currently, \demoreward is restricted to binary success detection. Future work could extend the framework to discrete or continuous reward signals \citep{RoboDopamine, RoboReward} and even preference-based annotations \citep{RL-VLM-F}, for example by optimizing correlation between reward predictions and temporal progress in demonstrations. Finally, \demoreward relies on an initial language instruction. An interesting direction for future research is reward modeling without any human language instructions, where a \vlm first infers a task description directly from demonstration videos and subsequently refines it through prompt optimization.

\appendix

\bibliography{main}
\bibliographystyle{rlj}

\beginSupplementaryMaterials

\section{Implementation Details}
\label{supp:impl_details}

\subsection{\demoreward}

\subsubsection{Hyperparameters}
\label{supp:hps}

Hyperparameters are shown per suite, with shared hyperparameters displayed in the center.

\begin{center}
\begin{tabular}{ l c c c}
\hline 
\textbf{Name} &   & \textbf{Value} & \\
 &  MetaWorld & RoboMimic & Real \\
\hline  
$\lambda$ &  & 0.01 & \\
Optimization steps $I$ &  & 100 &  \\
$m_\mathrm{critic}$ model &  \multicolumn{3}{c}{Qwen3-8B-VL}  \\
$m_\mathrm{meta}$ models &  \multicolumn{3}{c}{Qwen3-8B-VL \&  Qwen3-32B-VL}  \\
Optimizations $K$ per $m_\mathrm{meta}$ &  & 5 &  \\
$\vert\mathcal{D}\vert$ & 3 & 3 & 10 \\
image resolution \vlm inputs & \multicolumn{3}{c}{224 $\times$ 224 } \\
policy inputs & image & image & image \& position \\
image resolution policy inputs & 84$\times$84 & 96$\times$96 & 124$\times$124   \\
max episode steps & 100 & 200 (\texttt{Can}) / 300 (\texttt{Square}) & 30   \\

\hline
\end{tabular}    
\end{center}

\newpage

\subsubsection{Meta-Critic}
\label{supp:meta-critic}
The meta-critic uses the following system prompt:

\begin{tcolorbox}
\ttfamily

You are the Meta-Critic, an expert at writing precise, unambiguous text instructions for another vision-language model ("the Critic") that must decide whether a simulated robot has completed a task in a given video.

The Critic receives:
\begin{itemize}
    \item A short sequence of RGB frames (a sparsely sampled video) from a single episode, ordered in time from the start to the current state (roughly 5–10 frames, every Nth frame).
    \item Your instruction text.
\end{itemize}

The Critic outputs task completion score as a single, binary character: "1" if the task is completed in the last frame; "0" otherwise. Different formatting is not allowed.

Your goal is to rewrite the Critic's instruction so that its decisions align with the binary ground-truth labels of success/no success.

Follow these principles at all times:

\begin{itemize}
\item Define success strictly by clear visual evidence in the frames of the sequence.
\item Assume the video is sparsely sampled: avoid criteria that depend on seeing every moment of motion or exact trajectories.
\item State the task goal succinctly, then specify a minimal checklist of visual cues required for success.
\item Include explicit failure/ambiguity rules.
\item Specify what to ignore.
\item When needed, explicitly say how to compare frames.
\item Use imperative voice, and refer to objects as they appear visually.
\item Format your response exactly as follows without other additions or changes in formatting:
\end{itemize}

Reasoning: <3-5 sentences  about the ambiguities or mistakes of the current prompt that explain the failure cases.>

Final Instruction: <updated and final instruction text>

\end{tcolorbox}

At each optimization iteration, we provide the following inputs to the meta-critic for optimizing the prompt:
\begin{itemize}
    \item the current best prompt $p$ 
    \item a randomly sampled demonstration episode $\mathbf{o}_{1:T}, r_{1:T}$
    \item the confusion matrix of $p$ when evaluating $m_\mathrm{critic}$ on dataset $\mathcal{D}$, \ie TPR($p$), TNR($p$), FPR($p$), FNR($p$)
    \item 5 random subsampled episodes (5 frames) for which $m_\mathrm{critic}$ made a mistake with $p$
\end{itemize}
The meta-critic is prompted for an improved prompt $p'$ as follows:

\begin{tcolorbox}
\ttfamily
Rewrite the instruction that will be given to a separate AI model (the Critic). The Critic receives a short sequence of RGB frames from a single episode and must decide whether a simulated robot has completed the task. Your goal is to modify the instruction so that the Critic's binary decisions (1 = success, 0 = not successful) match the ground-truth labels as closely as possible.

As a reference, you are given a demonstration of a successful task execution: a sequence of frames from a single episode, each followed by a binary ground-truth label indicating whether the task is already completed in that particular frame. Use this reference, together with the performance summary to refine the instruction.

Ground-truth reference demonstration (each is an image followed by its ground-truth labels):
<$\mathbf{o}_1$> <$r_1$>
$\dots$
<$\mathbf{o}_T$> <$r_T$>

Current Critic instruction: <$p$>

Performance summary of the current Critic instruction over evaluation episodes:

True Positive Rate: <TPR($p$)>

True Negative Rate: <TNR($p$)>

False Positive Rate: <TPR($p$)>

False Negative Rate: <TNR($p$)>

Here are exemplar videos which the Critic judged incorrectly:

<$\mathbf{o}^i_1$> <$r^i_1$>
$\dots$
<$\mathbf{o}^i_K$> <$r^i_K$>
Incorrect Critic output:  <$m_\mathrm{critic}(\mathbf{o}^i_{1:K}, p)$

\dots

<$\mathbf{o}^j_1$> <$r^j_1$>
$\dots$
<$\mathbf{o}^i_L$> <$r^i_L$>
Incorrect Critic output:  <$m_\mathrm{critic}(\mathbf{o}^j_{1:L}, p)$

First provide the reasoning (3–5 sentences), then provide the instruction — both exactly as required by the output format.

\end{tcolorbox}

 We apply the same prompt for all tasks, but we dismiss the word ``simulated'' when applying \demoreward to real robots. 

For the meta-critic, we use an exploratory decoding scheme with stochastic sampling (temperature=1.3, top-p =0.97, top-k=50) and an increased token budget to encourage diverse and creative prompt refinements.

We employ both Qwen3-VL-8B-Instruct as well as Qwen/Qwen3-VL-32B-Instruct as meta-critics. Each of the two models is optimized $K=5$ times, from different random initializations for $I=100$ optimization steps.

\subsubsection{\vlmsd Critic}

\label{supp:vlmsd}

Our \vlmsd critic is based on the implementations of \cite{WorldGym} and \cite{WorldGymnast}. 
Unlike the implementation of \cite{WorldGym}, which uses GPT-4, we use \texttt{Qwen3-VL-8B-Instruct} as our \vlm. 
We slightly modify the original prompts to match our setup. 
Since \cite{WorldGym} assigns rewards to video diffusion rollouts, we remove all references to video diffusion. 
To increase throughput, we uniformly subsample video inputs to five frames. 
In addition, our \vlmsd outputs only the binary reward and does not generate reasoning traces. 
Ablation studies in \supp{supp:ablation} show that these modifications improve performance in our setup. At the same time our modification increase inference throughput by approximately a factor of ten in wall-clock time.

We use the same system prompt for all tasks. We only dismiss the word ``simulated'' when applying \demoreward to real robots. 

\begin{tcolorbox}
\ttfamily
You are an expert roboticist tasked to decide whether a simulated robot has completed a given task using a short video and a task instruction.

Output format:

Return a single character with no extra text:

"1" if the task is completed in this frame;

"0" otherwise.

Do not explain your answer.

\end{tcolorbox}

We use the following prompt to evaluate an episode:

\begin{tcolorbox}
\ttfamily
Here is a sequence of frames showing a robot policy attempting to solve a task. I need your help determining whether the policy is successful.

<$\mathbf{o}_1$>
$\dots$
<$\mathbf{o}_T$> 

Instruction: {$p$}

Output EXACTLY a single character, either 0 or 1, to denote task completion. Use 1 if the task is completed; 0 otherwise. Use no other symbols or formatting.

\end{tcolorbox}

When prompting the \vlmsd critic, we disable sampling for maximal accuracy.

\subsection{\gvl}

While we closely follow the prompt instructions provided in the original paper by \cite{GVL}, pre-tests showed that the \textsc{Vlm} sometimes didn't adhere to the desired output pattern when prompted to output a task completion score \emph{AND} a textual reasoning for this score. We assume that this happens because of a very long context length. 

We make two minor changes in order to reduce context length: 1) We uniformly subsample the demonstration video to 10 frames and subsample the policy video to 5 frames. 2) We prompt the \textsc{Vlm} to only output the reward not a video description. On an offline dataset we verified that this increased reward prediction accuracy. 

We use the following system prompt for \gvl:

\begin{tcolorbox}
\ttfamily
You are an expert roboticist tasked with predicting task completion percentages.

The task completion percentage must be between 0 and 100, where:
- 0$\%$ corresponds to the initial state
- 100$\%$ corresponds to full task completion

Frames may be presented in random order. Do not assume any temporal ordering.
Estimate completion based only on the visual state of each frame.

For each frame, output strictly:
"Frame {i}: Task Completion Percentages:{}$\%$"
Do not include any additional text.
\end{tcolorbox}

We use the following prompt for a given demonstration $(\mathbf{o}^d_1, \dots, \mathbf{o}^d_T)$ and policy video $(\mathbf{o}_1, \dots, \mathbf{o}_T)$:

\begin{tcolorbox}
\ttfamily
You are an expert roboticist tasked with predicting task completion percentages for a robot performing the following task: <task\_description>. The task completion percentages are between 0 and 100, where 100 corresponds to full task completion. 

We provide several examples of the robot performing the task at various stages and their corresponding task completion percentages. Note that these frames are in random order, so please pay attention to the individual frames when reasoning about task completion percentage.

Example frames (with known completion):

Frame 1: <$\mathbf{o}^d_i$> 

Task completion: <$i\%$>

$\dots$

Frame 10: <$\mathbf{o}^d_j$> 

Task completion:<$j\%$>

Now consider a new episode.

Initial frame of this episode:
<$\mathbf{o}^d_i$>

In the initial robot scene, the task completion percentage is 0.

Now, estimate task completion for the task: <task\_description>. Output the task completion percentage for the following frames that are presented in random order. For each frame, format your response as follow: Frame {{i}}: Task Completion Percentages:{{}}$\%$

Frame 1:
<$\mathbf{o}_h$>

$\dots$

Frame <t>:
<$\mathbf{o}_T$>

$\dots$

Frame 5:
<$\mathbf{o}_k$>
\end{tcolorbox}

We only use the output for time step $T$ as the reward for this episode. We disable sampling in the \vlm for maximal accuracy.

\subsection{RoboCLIP}
We closely followed the RoboCLIP setup \citep{RoboClip} by subsampling each video to 32 frames and resizing frames to $250 \times 250$, matching the preprocessing used to pretrain the S3D backbone \citep{S3D} on the HowTo100M dataset \citep{HowTo100M}. However, in preliminary experiments we observed that the raw dot-product similarity used by RoboCLIP exhibits substantial variation in magnitude across successful demonstrations from the same tasks. The reward values differed by up to three orders of magnitude for different videos of the same MetaWorld task. 

Since \textsc{Ibrl} requires rewards in [0,1], as it was pre-trained with binary task rewards and mixes value estimates during RL updates, we apply a fixed task-dependent normalization. For each task, we use the three available demonstrations and generate all possible 32-frame subclips. We compute dot-product similarities between the target demonstration embedding and all subclip embeddings to obtain a calibration distribution. We then estimate the 5th and 95th percentiles of this distribution and linearly rescale rewards to [0,1], clipping values outside this range. This transformation preserves the ranking induced by RoboCLIP while ensuring compatibility with our policy learning.

\subsection{RoboReward}

RoboReward \citep{RoboReward} is a pre-trained \texttt{Qwen3-VL-8B-Instruct} model fine-tuned on a large-scale robotics dataset to serve as a reward model. 
To ensure it works as intended, we use the same prompt that it was trained on:

\begin{tcolorbox}
\ttfamily
Given the task, assign a discrete progress score reward (1,2,3,4,5) for the robot in the video in the format: ANSWER: <score>

Rubric for end-of-episode progress (judge only the final state without time limits):

1 - No Success: Final state shows no goal-relevant change for the command.

2 - Minimal Progress: Final state shows a small but insufficient change toward the goal.

3 - Partial Completion: The final state shows good progress toward the goal but violates more than one requirement or a major requirement.

4 - Near Completion: Final state is correct in region and intent but misses a single minor requirement.

5 - Perfect Completion: Final state satisfies all requirements.

Task: <task\_description>
\end{tcolorbox}

Since RoboReward was fine-tuned using full videos, we do not subsample the video inputs to maximally align it with its fine-tuning setup.  We normalize RoboReward's output to $[0, 1]$ to match the reward scale of the binary rewards in the demonstration data.
We experimentally evaluate this design choice and compare it to unnormalized or binarized versions in \supp{supp:roboreward}.

\subsection{\textsc{Icl} \vlmsd}

In \supp{supp:ablation} we test an in-context learning (\textsc{Icl}) version for our \vlm success detector. For this ablation we use the following system prompt:

\begin{tcolorbox}
\ttfamily
You are an expert roboticist tasked to decide whether a simulated robot has completed a given task using two short videos and a task instruction.

The first video shows a demonstration of successful task completion.

The second video shows a robot policy attempting the task.

Output format:

Return a single character with no extra text:

"1" if the task is completed in the second video;

"0" otherwise.

Do not explain your answer.

\end{tcolorbox}

We randomly sample a demonstration $\mathbf{o}^d_{1:T}$ for \textsc{Icl} (sub-sampled to 10 frames) and provide it together with the task behavior video $\mathbf{o}_{1:T}$ (sub-sampled to 5 frames):

\begin{tcolorbox}
\ttfamily
Here are two short videos related to a robot task.

The FIRST video is a demonstration of a robot successfully completing the task.

The SECOND video shows a robot policy attempting the same task.

<$\mathbf{o}^d_1$>
$\dots$
<$\mathbf{o}^d_T$> 

<$\mathbf{o}_1$>
$\dots$
<$\mathbf{o}_T$> 

Task description: <task\_description>.

Using the demonstration video as a reference for what constitutes success, determine whether the robot in the SECOND video successfully completes the task.

Output EXACTLY a single character, either 0 or 1, to denote task completion. Use 1 if the task is completed; 0 otherwise. Use no other symbols or formatting.

\end{tcolorbox}

\section{Experiment Details}

\subsection{Simulated Experiments}
\label{supp:sim_details}

For all simulated experiments, we follow the setup of \textsc{Ibrl} \citep{IBRL} and use their implementation of \textsc{Ibrl} and \textsc{Rlpd}. 
We modify only one aspect: rewards are provided exclusively at the end of each episode, consistent with prior work on \vlm-based reward models \citep{RoboClip, RoboReward}. 
In particular, we disable early episode termination and assign rewards at the end of truncated episodes. 
Disabling early termination is necessary to prevent task success signals from leaking through episode termination conditions.

In all experiments, the \vlm receives higher-resolution images (224 $\times$ 224) than the policies to enable more accurate visual reasoning (see \supp{supp:hps}). 
Following \cite{IBRL}, we use a third-person camera for \texttt{NutAssembly-Square} and a gripper-mounted camera for \texttt{PickPlace-Can}. 
For reward evaluation, however, the \vlm always receives third-person views to simplify visual assessment.

For MetaWorld we took the generic task descriptions from the benchmark website: \url{https://metaworld.farama.org/benchmark/task_descriptions/}. 
For RoboMimic we use the official task descriptions from the RoboSuite paper \citep{Robosuite}.

\paragraph{Motivating Example.}
For the motivating example in Section~\ref{sec:reward_hacking}, we use a slightly modified setup to more clearly illustrate the effects of reward hacking. 
In this case, the \vlm is queried at every time step to produce a reward $\hat{r}_t$. 
A \textsc{Ibrl} policy is then trained using these synthetic rewards. 
While this setup clearly highlights the impact of false positive rewards, it is computationally infeasible for real-world robotics applications due to the high inference cost of querying the \vlm at every transition.

\subsection{Real Robot Experiments}
\label{supp:robot}
For our real-world evaluation, we deploy our method on a Franka Research 3 manipulator. We build upon the RLInf framework \citep{RLInf}, adopting both their ResNet-based pre-trained visual encoder and their implementation of \textsc{Rlpd}. However, we introduce key architectural modifications to the teleoperation and vision pipelines. Specifically, we integrate a ZED 2i stereo camera through ROS2 and employ the GELLO framework \citep{gello} for teleoperation, using its native Franka controller and joint state publisher.

\paragraph{Data processing} To seamlessly interface our hardware setup with the RLInf training pipeline, we implemented a custom data-logging module tailored to match their expected data format. Since precise temporal alignment between visual observations and robot proprioception is critical for RL, we enforce strict cross-machine clock synchronization using Chrony (NTP). We combine this with ROS2 time synchronization policies to tightly couple incoming image frames with their corresponding robot state messages. Following \cite{RLInf}, the recorded end-effector poses are mapped into a relative end-effector coordinate frame. The continuous action space is defined by the position deltas between consecutive robot states, which are subsequently normalized to align with the action scaling of the rollout environment.

\paragraph{\texttt{Lid-On-Pot}} This task is inspired by the two insertion tasks used in RLInf \citep{RLInf}, but adapted for our robot and a kitchen setup. 
Our robot holds the lid and must move it ontop of the pot to close it. 
We measure ground-truth success based on the end effector position. Specifically a target pose ([0.541, -0.027, 0.101]) must be reached within a certain threshold ([0.025, 0.025, 0.035]). 
The observation space for the end-effector poses is symmetric around the target with 10cm in both x and y direction, and extends 10cm above the target in the z coordinate. We truncate episodes after 30 time steps and allocating rewards strictly at truncation.

\section{Additional Results}

\subsection{Example optimizations and annotations}
\label{supp:examples}

\begin{figure}[h]
    \centering
\begin{subfigure}{0.66\linewidth}
\centering
\includegraphics[width=\linewidth]{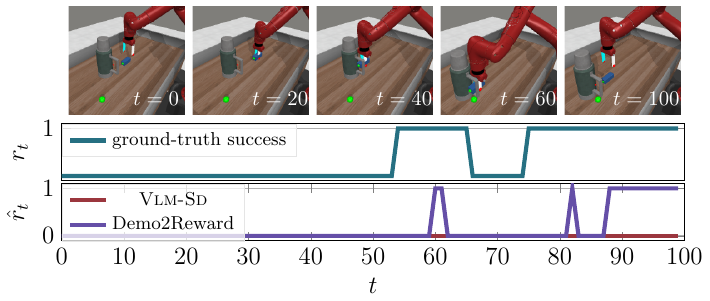}
\caption{ \textbf{Reward predictions for \texttt{Stick-Pull}}}
\end{subfigure}
\hfill
\begin{subfigure}{0.32\linewidth}
\centering
\includegraphics[width=\linewidth]{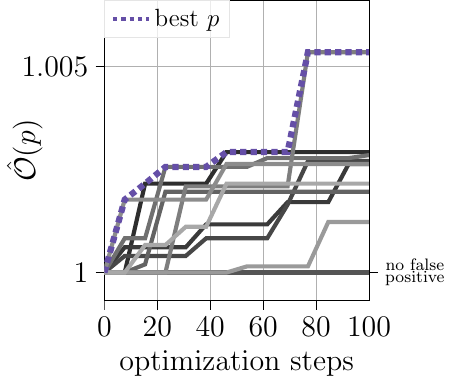}
\caption{$\hat{\mathcal{O}}$ for \texttt{Stick-Pull} }
\end{subfigure}
\begin{subfigure}{0.66\linewidth}
\centering
\includegraphics[width=\linewidth]{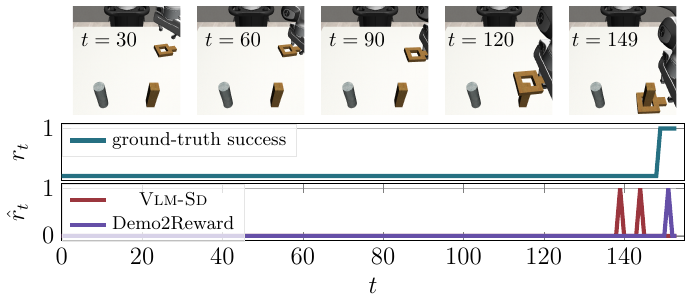}
\caption{ \textbf{Reward predictions for \texttt{AssembleNut-Square}}}
\end{subfigure}
\hfill
\begin{subfigure}{0.32\linewidth}
\centering
\includegraphics[width=\linewidth]{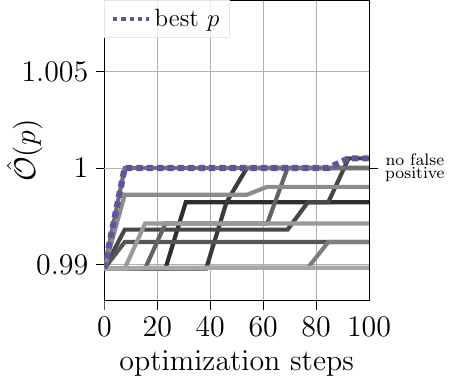}
\caption{$\hat{\mathcal{O}}$ for \texttt{Assemble-Square}}
\end{subfigure}
    \caption{\textbf{Prompt optimization}: Example reward prediction for \texttt{Stick-Pull}  \textbf{(a)} and \texttt{AssembleNut-Square}  \textbf{(c)} before (\textcolor{darkred}{red}) and after (\textcolor{newpurple}{purple}) \demoreward. Objective $\hat{\mathcal{O}}(p)$  (\eqn{eq:O}) for \texttt{Stick-Pull}  \textbf{(b)} and \texttt{AssembleNut-Square}  \textbf{(d)}  over optimization steps. Each gray line shows one of $K=10$ independent optimization runs with the dashed \textcolor{newpurple}{purple} marking the best. The $y$-axis is displayed in log-scale.} 
\end{figure}

\begin{minipage}{0.4\linewidth}
\centering
\captionof{table}{Confusion matrix for $\mathcal{D}_\mathrm{sub}$ in \texttt{Assemble-Nut} \textit{before} optimization}
\begin{tabular}{c|cc}
 & \multicolumn{2}{c}{\textbf{Pred. Reward} $\hat{r}_t$} \\
\textbf{Success} $r_t$ & 1 & 0 \\ \hline
1 & 0.6$\%$ & 99.4$\%$ \\
0 &  5.3$\%$ & 94.7$\%$ \\
\end{tabular}
\end{minipage}
\hfill
\begin{minipage}{0.4\linewidth}
\centering
\captionof{table}{Confusion matrix for $\mathcal{D}_\mathrm{sub}$ in \texttt{Assemble-Nut} \textit{after} optimization}
\begin{tabular}{c|cc}
 & \multicolumn{2}{c}{\textbf{Pred. Reward} $\hat{r}_t$} \\
\textbf{Success} $r_t$ & 1 & 0 \\ \hline
1 & 14.9$\%$ & 85.1$\%$ \\
0 & 0$\%$ & 100$\%$ \\
\end{tabular}
\end{minipage}

\begin{minipage}{0.4\linewidth}
\centering
\captionof{table}{Confusion matrix for $\mathcal{D}_\mathrm{sub}$ in \texttt{Stick-Pull} \textit{before} optimization}
\begin{tabular}{c|cc}
 & \multicolumn{2}{c}{\textbf{Pred. Reward} $\hat{r}_t$} \\
\textbf{Success} $r_t$ & 1 & 0 \\ \hline
1 & 0$\%$ & 100$\%$ \\
0 & 0$\%$ & 100$\%$ \\
\end{tabular}
\end{minipage}
\hfill
\begin{minipage}{0.4\linewidth}
\centering
\captionof{table}{Confusion matrix for $\mathcal{D}_\mathrm{sub}$ in \texttt{Stick-Pull} \textit{after} optimization}
\begin{tabular}{c|cc}
 & \multicolumn{2}{c}{\textbf{Pred. Reward} $\hat{r}_t$} \\
\textbf{Success} $r_t$ & 1 & 0 \\ \hline
1 & 41.3$\%$ & 58.7$\%$ \\
0 & 0$\%$ & 100$\%$ \\
\end{tabular}
\end{minipage}

\begin{minipage}{0.4\linewidth}
\centering
\captionof{table}{Confusion matrix for $\mathcal{D}_\mathrm{sub}$ in \texttt{NutAssemble-Square} \textit{before} optimization}
\begin{tabular}{c|cc}
 & \multicolumn{2}{c}{\textbf{Pred. Reward} $\hat{r}_t$} \\
\textbf{Success} $r_t$ & 1 & 0 \\ \hline
1 & 33.3$\%$ & 66.7$\%$ \\
0 & 1.4$\%$ & 98.6$\%$ \\
\end{tabular}
\end{minipage}
\hfill
\begin{minipage}{0.4\linewidth}
\centering
\captionof{table}{Confusion matrix for $\mathcal{D}_\mathrm{sub}$ in \texttt{NutAssemble-Square} \textit{after} optimization}
\begin{tabular}{c|cc}
 & \multicolumn{2}{c}{\textbf{Pred. Reward} $\hat{r}_t$} \\
\textbf{Success} $r_t$ & 1 & 0 \\ \hline
1 & 6.7$\%$ & 93.3$\%$ \\
0 & 0$\%$ & 100$\%$ \\
\end{tabular}
\end{minipage}

\subsection{Discovered Instructions}
\label{supp:prompts}
Here we provide all instructions discovered by \demoreward.

\subsubsection{MetaWorld}
The following prompt were used for our main experiments in \sec{sec:res1}.

\textbf{\texttt{Assemble-Nut}}

Before \demoreward:
\begin{tcolorbox}
\ttfamily
  Evaluate the behavior of this simulated robot. The robot was instructed with the task: Pick up a nut and place it onto a peg.
\end{tcolorbox}

After \demoreward:
\begin{tcolorbox}[colback=newpurple!20, colframe=newpurple]
\ttfamily
  In the final frame, determine if the robot has successfully completed the task by visually verifying that the nut (a gray circular object) is fully seated on top of the peg (a red cylindrical object), with no visible separation between them. Output “1” if the nut is completely resting on the peg; otherwise, output “0”. Ignore intermediate actions or partial placements. Do not consider motion or rotation as failure — success is determined by final positional alignment.
\end{tcolorbox}

\textbf{\texttt{Box-Close}}

Before \demoreward:
\begin{tcolorbox}
\ttfamily
  Evaluate the behavior of this simulated robot. The robot was instructed with the task: Grasp the cover and close the box with it.
\end{tcolorbox}

After \demoreward:

\begin{tcolorbox}[colback=newpurple!20, colframe=newpurple]
\ttfamily
  In the final frame, determine if the robot has successfully completed the task "grasp the cover and close the box with it" by checking whether the cover is visibly grasped by the robot’s end-effector and placed on top of the box such that it overlaps with the box’s top surface; ignore incomplete alignment or small gaps. If this condition is met, output "1"; otherwise, output "0".
\end{tcolorbox}

\textbf{\texttt{Coffee-Push}}

Before \demoreward:
\begin{tcolorbox}
\ttfamily
  Evaluate the behavior of this simulated robot. The robot was instructed with the task: Push a mug under a coffee machine.
\end{tcolorbox}

After \demoreward:

\begin{tcolorbox}[colback=newpurple!20, colframe=newpurple]
\ttfamily
  The robot's task is to push a white mug under a red coffee machine. Evaluate the final frame: if the white mug is fully under the coffee machine (no part of it is visible on the table surface) and there is a small green dot visible near the coffee machine’s spout (indicating correct placement), output "1". If the mug is still visible on the table, not positioned under the machine, or the green dot is absent, output "0". Ignore the robot’s arm position or motion. Do not consider partial positioning or intermediate contact as success. Success is defined solely by the mug’s final position and the presence of the green dot.
\end{tcolorbox}

\textbf{\texttt{Stick-Pull}}

Before \demoreward:
\begin{tcolorbox}
\ttfamily
  Evaluate the behavior of this simulated robot. The robot was instructed with the task: Grasp a stick and pull a box with the stick.
\end{tcolorbox}

After \demoreward:

\begin{tcolorbox}[colback=newpurple!20, colframe=newpurple]
\ttfamily
  Output "1" only if, in the final frame, the red stick is visibly held by the robot’s gripper and is in direct contact with the multicolored box, and the box is positioned close to the green ball (within touching distance or adjacent). Ignore motion or intermediate frames. If the box is not near the green ball, or the stick is not clearly contacting the box, or the stick is not gripped, output "0". The gripper must be visibly closed around the stick, and the stick must appear to be touching the side or top of the multicolored box.
\end{tcolorbox}

\subsubsection{MetaWorld Alternativ}
For our robustness analysis in \sec{sec:res2} we generated two addittional prompts per task.

\textbf{\texttt{Box-Close}-2}

\begin{tcolorbox}[colback=newpurple!20, colframe=newpurple]
\ttfamily
  The robot's task is to grasp the cover and close the box with it. The task is completed in the final frame if and only if the wooden box is fully closed (no open sides visible, lid fully seated) and the box is visibly lifted off the table (with a clear gap between the bottom of the box and the table surface). If both conditions are met, output 1. If the box is still in contact with the table or visibly open, output 0. Do not require the green dot to be fully covered—its partial visibility does not invalidate closure. Ignore intermediate steps such as grasping or aligning the cover. When in doubt, if the box appears closed and lifted, output 1.
\end{tcolorbox}

\textbf{\texttt{Box-Close}-3}

\begin{tcolorbox}[colback=newpurple!20, colframe=newpurple]
\ttfamily
  Output "1" if in the last frame the wooden box is lifted by the robot's gripper and is visibly not resting on the table (even if close to the surface), and the green dot is visible on the top face of the box (regardless of exact centering or partial occlusion). Output "0" only if the box is touching the table or the green dot is not on the top surface (e.g., invisible, on side, or absent). Ignore motion blur, intermediate steps, and minor changes in arm pose. Focus exclusively on the position of the box and visibility of the green dot in the final frame.
\end{tcolorbox}

\textbf{\texttt{Coffee-Push}-2}

\begin{tcolorbox}[colback=newpurple!20, colframe=newpurple]
\ttfamily
  In the final frame, if the mug is visibly placed under the spout of the coffee machine — meaning the top of the mug is within the projected area of the spout’s downward cone (even if slightly off-center or tilted) — output 1. Otherwise, output 0. Ignore robot motion or intermediate positioning — only evaluate whether the mug is located under the spout’s projection in the final frame.
\end{tcolorbox}

\textbf{\texttt{Coffee-Push}-3}

\begin{tcolorbox}[colback=newpurple!20, colframe=newpurple]
\ttfamily
 In the final frame, determine if the robot has completed the task by visually confirming that the white mug is positioned within the dispensing area of the red coffee machine — meaning the mug must be visibly beneath the spout, even if slightly offset or not perfectly centered. Output "1" if the mug is visibly under the spout (any portion of the mug overlaps the spout's projection); output "0" only if no part of the mug is under the spout at all. Ignore motion or trajectory in preceding frames — only the final frame matters. Do not consider the gripper state or its orientation in your decision.
\end{tcolorbox}

\subsubsection{RoboMimic}

\textbf{\texttt{PickPlace-Can}}

Before \demoreward:
\begin{tcolorbox}
\ttfamily
  Evaluate the behavior of this simulated robot. The robot was instructed with the task: A can is placed in a bin in front of a single robot arm. There are four containers next to the bin. The robot must place the can into its corresponding container.
\end{tcolorbox}

After \demoreward:

\begin{tcolorbox}[colback=newpurple!20, colframe=newpurple]
\ttfamily
  In the final frame, output "1" only if the red can is visually fully and stably seated inside a compartment of the container AND the robot’s arm and gripper are stationary, disengaged from the can, and not in motion — otherwise output "0".
\end{tcolorbox}

\textbf{\texttt{AssembleNut-Square}}

Before \demoreward:
\begin{tcolorbox}
\ttfamily
  Evaluate the behavior of this simulated robot. The robot was instructed with the task: Two colored pegs (one square and one round) are mounted on the tabletop, and a square nut is placed on the table in front of a single robot arm. The robot must fit the square nut onto the square peg.
\end{tcolorbox}

After \demoreward:

\begin{tcolorbox}[colback=newpurple!20, colframe=newpurple]
\ttfamily
  The robot must fit the brown square nut onto the brown square peg. Success is achieved in the final frame if the brown square nut is fully seated on top of the brown square peg with no visible gaps, and the gripper is not in contact with or holding the nut. The nut and peg must be in stable physical contact with no motion. If the nut is floating above, partially inserted, or the gripper is touching the nut, output "0". Ignore slight rotation or minor misalignment if the nut is visibly seated and stable. Output "1" only if the nut is clearly seated and not being held.
\end{tcolorbox}

\subsubsection{Real Robot}

\textbf{\texttt{Lid-On-Pot}}

Before \demoreward:
\begin{tcolorbox}
\ttfamily
  Evaluate the behavior of this simulated robot. The robot was instructed with the task: Cover the pot with the lid.
\end{tcolorbox}

After \demoreward:

\begin{tcolorbox}[colback=newpurple!20, colframe=newpurple]
\ttfamily
  The robot's task is to place the lid fully on the pot. The task is complete if, in the final frame, the lid is resting stably on top of the pot, covering its opening without being held by the gripper. The lid must appear seated and not tilted or suspended. If the gripper is still touching the lid or is positioned directly over it in a holding pose, the task is not complete. If the lid is clearly and completely on the pot with no visible motion or support from the gripper, output "1". Otherwise, output "0". Ignore intermediate states or near-completion poses not clearly showing success.
\end{tcolorbox}

\subsection{Ablation of \vlmsd}
\label{supp:ablation}

\begin{figure}
    \centering
    {\scriptsize
      \textcolor{darkdarkred}{\rule[2pt]{20pt}{2pt}} Full Video \quad
      \textcolor{lightdarkred}{\rule[2pt]{20pt}{2pt}} Final Frame \quad
      \textcolor{magenta}{\rule[2pt]{20pt}{2pt}} Descr. \quad
      \textcolor{darkwmgreen}{\rule[2pt]{20pt}{2pt}} \textsc{Icl} \quad
      \textcolor{darkbrown}{\rule[2pt]{20pt}{2pt}} Qwen32 \quad
      \textcolor{darkred}{\rule[2pt]{20pt}{2pt}} base
    }\\
    \vspace{.3em}
\begin{subfigure}{0.45\linewidth}
\centering
\includegraphics[width=\linewidth]{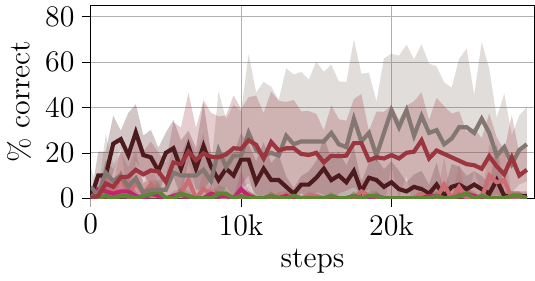}
\caption{\texttt{Box-Close}}
\end{subfigure}
\hfill
\begin{subfigure}{0.45\linewidth}
\centering
\includegraphics[width=\linewidth]{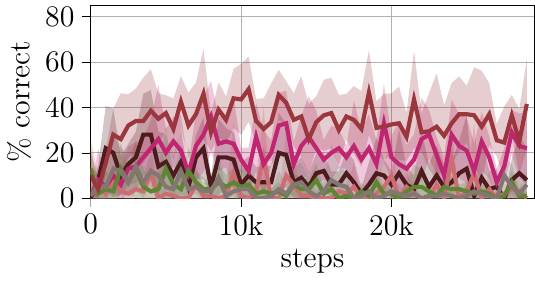}
\caption{\texttt{Coffee-Push}}
\end{subfigure}
    \caption{\textbf{Ablations of \vlm success detector} when used for downstream policy learning with \textsc{Ibrl} in MetaWorld. We compare full video inputs (Full Video), using only a single, final frame (Final Frame), reasoning descriptions (Descr.), In-Context Learning (\textsc{Icl}), a larger \vlm backbone (Qwen32) to our base \vlmsd. Shaded area shows standard deviation (5 seeds). \label{fig:ablations}}
\end{figure}

Our \vlm success detector (\vlmsd) is based on the prompting scheme of \cite{WorldGym}, with two main modifications: (1) uniform subsampling of input videos to five frames, and (2) restricting the \vlm to output only a binary reward rather than a textual explanation. Here, we evaluate the impact of these design choices. 
First we compare against using the full video and to even stronger subsampling, providing only the last frame of the video. 
Next, we compare against an ablation additionally outputting reasoning for each reward.
We also compare different \vlm sizes, namely \texttt{Qwen3-VL-8B} and \texttt{Qwen3-VL-32B}. Finally, motivated by \cite{GVL}, we analyze whether providing an in-context demonstration (\textsc{ICL}) improves reward prediction. All ablations are evaluated on \texttt{Coffee-Push} and \texttt{Box-Close}, two representative MetaWorld tasks.

Figure~\ref{fig:ablations} reports the resulting success rates when using the rewards generated by these models for training an \textsc{Ibrl} policy. For both tasks, using the larger \vlm does not substantially improve performance or even degrades it. This is consistent with findings from the RoboReward benchmark \citep{RoboReward}, where \texttt{Qwen3-VL-8B} also outperformed \texttt{Qwen3-VL-32B}. Generating additional reasoning (Descr.), using in-context learning (\textsc{Icl}), or providing the full video or final frame instead of subsampled frames (Full Video) all lead to worse downstream policy performance. We hypothesize that longer contexts increase task complexity for the \vlm, whereas focusing on a small set of informative frames and a single binary output improves reliability.
Based on these results, we adopt \texttt{Qwen3-VL-8B} with five-frame subsampling and binary-only outputs as our default \vlmsd configuration.

\subsection{RoboReward ablations}
\label{supp:roboreward}

Rewards from RoboReward \citep{RoboReward} do not lead to meaningful policies in our experiments with \textsc{Ibrl}. 
This is surprising, as successful downstream policy learning was reported by the authors. 
Since our only modification to their approach was reward normalization, we experimentally analyze the effect of this design choice. 
In addition, as \textsc{Ibrl} was originally designed for binary rewards, we consider a second ablation in which we binarize RoboReward outputs by treating near completion (4) and perfect completion (5) as success ($\hat{r}_t = 1$), and all other outputs as failure ($\hat{r}_t = 0$).

\begin{figure}[b]
    \centering
    {\scriptsize
      \textcolor{darkdarkwmgreen}{\rule[2pt]{20pt}{2pt}} {unnorm.} \quad
      \textcolor{darkwmgreen}{\rule[2pt]{20pt}{2pt}} {norm.} \quad
      \textcolor{lightdarkwmgreen}{\hdashrule[2pt]{20pt}{2pt}{2pt 1pt}} {binary} \quad
    }\\
    \vspace{.3em}
\begin{subfigure}{0.45\linewidth}
\centering
\includegraphics[width=\linewidth]{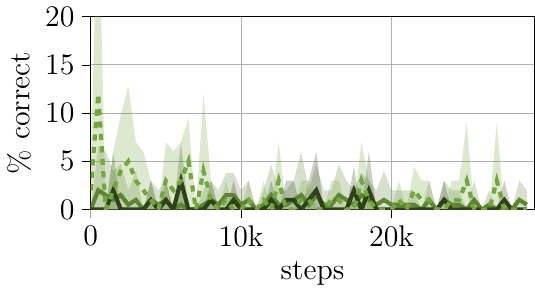}
\caption{\texttt{Box-Close}}
\end{subfigure}
\hfill
\begin{subfigure}{0.45\linewidth}
\centering
\includegraphics[width=\linewidth]{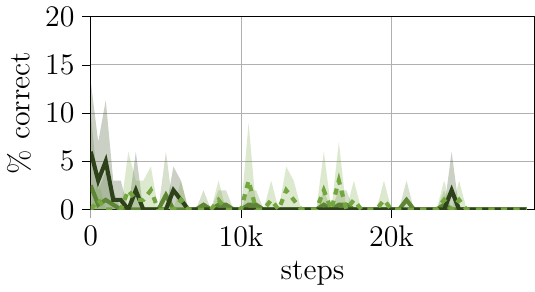}
\caption{\texttt{Coffee-Push}}
\end{subfigure}
    \caption{\textbf{Ablations of RoboReward} when used for downstream policy learning with \textsc{Ibrl} in MetaWorld. We compare unnormalized discrete rewards (1-5), to normalized rewards and to binary rewards. Shaded area shows standard deviation (5 seeds). \label{fig:roboreward_ablation}}
\end{figure}

\Fig{fig:roboreward_ablation} shows the resulting success rates when training \textsc{Ibrl} policies on two representative MetaWorld tasks, \texttt{Coffee-Push} and \texttt{Box-Close}, using these reward variants. 
None of the ablations achieve consistent task success. 
We therefore conclude that reward normalization is unlikely to be the primary cause of the observed performance degradation. 
Instead, the issue may stem from deeper factors, such as limited transfer between RoboReward’s training data of real-world robotics episodes and our simulation setup.

\end{document}